\documentclass[letterpaper, 12pt]{ewshm}

\usepackage[dvips,final]{graphicx}

\usepackage[table]{xcolor}
\usepackage{makecell}
\usepackage{hyperref}
\usepackage{booktabs}
\usepackage{enumerate}
\usepackage{color}
\usepackage{subfig}
\usepackage{caption}
\usepackage{amsfonts}
\usepackage{threeparttable}
\usepackage{amsmath}
\usepackage{amssymb}
\usepackage{times}
\usepackage{cite}
\usepackage{hhline}
\usepackage{compactbib}
\usepackage{graphicx}        
\usepackage{amsmath,amsfonts,amssymb}
\usepackage{geometry}
\usepackage{url}
\usepackage[labelsep=period]{caption}

\definecolor{halfgreen}{RGB}{0,128,0}
\definecolor{ahsred}{RGB}{192,0,0}

\renewcommand{\thefigure}{\arabic{figure}}
\renewcommand{\thetable}{\Roman{table}}

\newcommand{\beq}{\begin{equation}}
\newcommand{\eeq}{\end{equation}}
\newcommand{\bgqar}{\begin{eqnarray}}
\newcommand{\enqar}{\end{eqnarray}}
\newcommand{\bgqarn}{\begin{eqnarray*}}
\newcommand{\enqarn}{\end{eqnarray*}}
\newcommand{\bgary}{\begin{array}}
\newcommand{\enary}{\end{array}}

\long\def\symbolfootnote[#1]#2{\begingroup%
\def\thefootnote{\fnsymbol{footnote}}\footnote[#1]{#2}\endgroup}

\makeatletter
\renewcommand\@biblabel[1]{#1.}
\makeatother

\begin{document}


\vspace*{4.4cm}

\noindent Title: \textbf{From Actions to Kinesics: Extracting Human Psychological States through Bodily Movements}

\vspace{1.6cm}

\noindent
$
\begin{array}{ll}
\text{Authors}: 
& \text{Cheyu Lin}  \\ 
& \text{Katherine A. Flanigan} 
\end{array}
$

\newpage


\vspace*{60mm}

\noindent \uppercase{\textbf{ABSTRACT}} \vspace{12pt} 

Understanding the dynamic relationship between humans and the built environment is a key challenge in disciplines ranging from environmental psychology to reinforcement learning (RL). A central obstacle in modeling these interactions is the inability to capture human psychological states in a way that is both generalizable and privacy preserving. Traditional methods rely on theoretical models or questionnaires, which are limited in scope, static, and labor intensive. We present a kinesics recognition framework that infers the communicative functions of human activity---known as kinesics---directly from 3D skeleton joint data. Combining a spatial-temporal graph convolutional network (ST-GCN) with a convolutional neural network (CNN), the framework leverages transfer learning to bypass the need for manually defined mappings between physical actions and psychological categories. The approach preserves user anonymity while uncovering latent structures in bodily movements that reflect cognitive and emotional states. Our results on the Dyadic User EngagemenT (DUET) dataset demonstrate that this method enables scalable, accurate, and human-centered modeling of behavior, offering a new pathway for enhancing RL-driven simulations of human-environment interaction.

\symbolfootnote[0]{\hspace*{-7mm} Cheyu Lin\textsuperscript{1}, Katherine A. Flanigan\textsuperscript{2}, Ph.D (Corresponding author). Email: \{cheyul\textsuperscript{1}, kflaniga\textsuperscript{2}\}@andrew.cmu.edu. Department of Civil and Environmental Engineering, Carnegie Mellon University, Pittsburgh, PA, USA.}


%

\vspace{12pt}
\noindent \uppercase{\textbf{INTRODUCTION}}  \vspace{12pt} 

In \textit{Space is the Machine}, Hiller argues that buildings are not merely shelters but also serve physical, spatial, and social functions that are deeply interconnected \cite{hillier2007space}. The physical structure defines and supports spatial configurations, while architectural style and details reflect both preferences and broader cultural meanings. These spatial layouts, in turn, shape accessibility, patterns of interaction, and overall social dynamics. The complex and reciprocal relationship between the built environment and human behavior is evident in various real-world contexts. For example, in nursing homes, the design and quality of physical spaces can influence residents’ psychological well-being and sense of home \cite{eijkelenboom2017architectural}. Similarly, the physical condition and organization of school buildings affect students’ health, cognitive development, and academic outcomes \cite{schneider2002school}. Such examples not only illustrate the mutual influence between humans and the urban fabric but also suggest that we can intentionally design physical, spatial, and social elements to promote social benefits \cite{doctorarastoo2023exploring,doctorarastooIFAC}. In practice, this has motivated both post-occupancy evaluation (POE) studies that assess how spaces are experienced and used after construction, and modeling approaches that aim to simulate such dynamics in advance. One prominent approach is agent-based modeling (ABM), which allows researchers to represent individual behavior and interaction within specific spatial layouts.

Agent-based modeling (ABM) captures the dynamics between interacting agents---such as humans and the built environment---and is particularly valued for its ability to encode decision-making processes within human agents \cite{crooks2011introduction}. This enables the simulation of human behaviors in hypothetical scenarios. However, agent rationale is often based on theoretical frameworks of planned behavior \cite{ajzen1991theory} or from manually collected data via questionnaires and expert opinions \cite{liang_how_2022}, both of which limit ABM's potential in key ways. First, theoretical models tend to oversimplify human decision making, neglecting the inherent heterogeneity of human behavior. Second, while questionnaires may better reflect cognitive complexity \cite{lin2024your}, they are time consuming, costly, and constrained by the limited number of processes humans can report or interpret. Moreover, such data only captures a snapshot of a person's psychological state, lacking the adaptability of continuously sensed information. These limitations hinder ABM's ability to represent human reasoning and preferences in a faithful and timely way. \textit{This underscores the need for sensing technologies that can infer psychological states while remaining human-centered.} Crucially, such technologies must prioritize user privacy---not all sensing approaches are appropriate for this task \cite{doctorarastoo2023modeling,linRoom}. We argue that effective solutions must extract rich, socially relevant signals that reflect mental reasoning, while explicitly excluding identifiable features to maintain trust between stakeholders and users.

To address the limitations of manually sourced cognitive data, we turn to an underutilized yet powerful modality for understanding human reasoning: body language. Humans naturally externalize thought processes through a range of channels, and among them, body language plays a critical role by conveying unspoken cues about an individual's mental and emotional state through movement \cite{sharan2022relative}. While these movements may initially seem unorganized, they are in fact structured by a psychological taxonomy developed by Ekman and Friesen \cite{ekman1969repertoire}. This taxonomy, known as kinesics, classifies body language into five functional categories: illustrators, regulators, affect displays, adaptors, and emblems. Each category provides a well-defined link between specific bodily actions and the meanings or intentions they express. For instance, a hug is classified as an affect display, signaling warmth and emotional closeness. This taxonomy enables a principled mapping between physical activity and underlying psychological state, setting the foundation for extracting cognitive insight from bodily behavior.

By pairing this taxonomy with human activity recognition (HAR) techniques---which use sensor data such as RGB video, depth maps, or 3D skeletal keypoints to identify actions---it becomes possible to infer the kinesic category of a given movement. However, this approach faces a major limitation: the sheer variety of human actions makes it infeasible to manually define mappings for every possible movement. Consequently, the generalizability of existing frameworks remains limited. To truly decode human reasoning from bodily movements, we must move beyond dictionary-based mappings and toward methods capable of learning a generalized translation between physical actions and their cognitive or affective significance.

To extract the kinesic function of human activities without relying on a predefined dictionary, we propose a kinesics recognition framework based on transfer learning. This framework leverages patterns inherently embedded in human activity data to infer the corresponding kinesic categories \cite{torrey2010transfer}. Specifically, our approach combines a frozen HAR model with a trainable convolutional neural network (CNN). The HAR component is implemented using a Spatial-Temporal Graph Convolutional Network (ST-GCN) \cite{yan2018spatial}, a skeleton-based model originally developed to classify activity types. Rather than using ST-GCN for activity recognition, we extract the latent features from its final hidden layer and use these as input to a CNN that classifies the activity's kinesic function. This architecture not only eliminates the need for a manually defined mapping between actions and kinesic categories, but also preserves user privacy by relying solely on 3D skeleton keypoints, which contain no identifiable features. The framework is applied to the Dyadic User EngagemenT (DUET) dataset \cite{huggingfaceAnonymousUploader1DUETDatasets}, a HAR dataset explicitly inspired by the psychological taxonomy of kinesics.

The remainder of this paper is structured as follows: Section 2 introduces the DUET dataset and the supporting kinesic taxonomy. Section 3 details the data preprocessing steps, the structure of the machine learning models, and the experimental results. The full implementation is publicly available on Hugging Face \cite{duetgithub}. We conclude in Section 4 with a discussion of key findings and directions for future work.

\vspace{24pt}

\noindent \uppercase{\textbf{DUET -- \underline{D}yadic \underline{U}ser \underline{E}ngagemen\underline{T} Dataset}} \vspace{12pt}

Integrating psychological theory with HAR is essential for interpreting behavioral coherence as an expression of human thought processes. In this section, we present the psychological taxonomy of kinesics and introduce the DUET dataset, a HAR dataset explicitly derived from this taxonomy.

\begin{figure}[!t] 
  \centering{\includegraphics[width=\columnwidth]{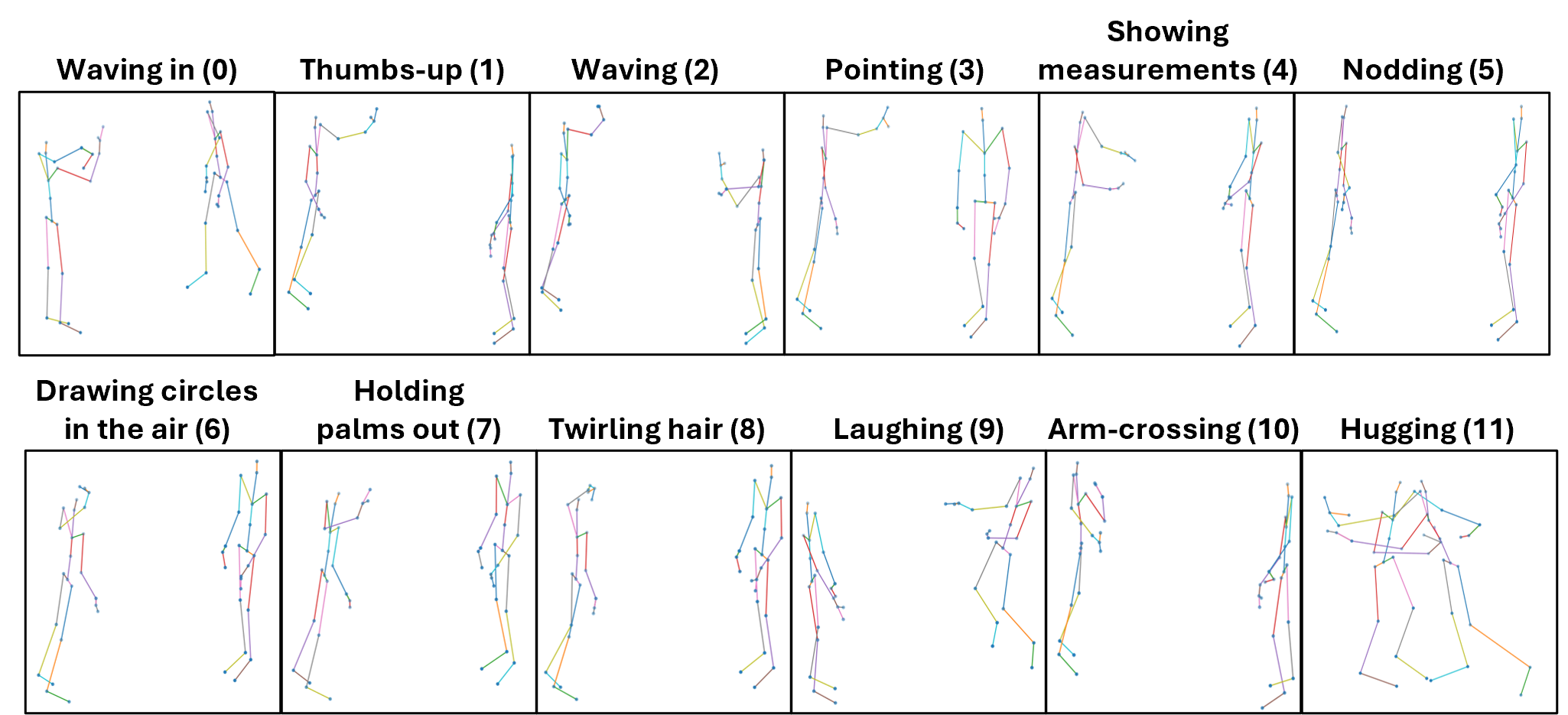}}
    \caption{\small Sample frames for each interaction (class label is denoted in parentheses).}
\label{fig:3d_keypoints}%
\end{figure}

DUET is a two-person (or ``dyadic''), multimodal HAR dataset. It contains 12 human activities spanning four sensing modalities, including RGB, depth, infrared, and 3D skeleton joints. The work in this paper only adopts 3D skeleton joints as the sensing modality, as shown in Figure \ref{fig:3d_keypoints}, to align with the privacy-preserving requirement of human-centered applications \cite{HM-SYNC}. The dataset was collected at an open indoor space, a confined indoor space, and an open outdoor space, allowing users to investigate the effects ambient environments impose on HAR algorithms. The 12 activities are adopted from the taxonomy of kinesics developed by Ekman and Friesen \cite{ekman1969repertoire}, which, to the best of our knowledge, is the only dataset that integrates HAR with a scientifically grounded psychology study. As described in Table \ref{table:lit5}, there are five categories in the taxonomy: emblems, illustrators, regulators, adaptors, and affect displays.

\begin{table*}[tb!]
\caption{\small Overview of Ekman and Friesen’s kinesic taxonomy \cite{ekman1969repertoire} with representative gestures from the DUET dataset.}
\vspace{-6pt}
\label{table:lit5}
\small
\centering
\begin{threeparttable}

\begin{tabular} {p{0.11\linewidth}>{\raggedright}p{0.62\linewidth}>{\raggedright\arraybackslash}p{0.18\linewidth}}
\toprule
{Taxonomy} & {Interaction Description} & {Example(s)} \\
\midrule
Emblems & Emblems are culturally specific gestures that have direct verbal equivalents. Their meaning is widely understood within a given culture but may differ significantly across cultural contexts. For example, a ``thumbs up'' signals approval in many Western cultures but is offensive in some Middle Eastern regions \cite{hartman_nonverbal_nodate}. & \textit{waving in (0)}, \textit{thumbs-up (1)}, \textit{hand waving (2)} \\ \addlinespace
Illustrators & Illustrators are gestures that complement and clarify spoken language by visually reinforcing verbal messages. They provide additional context to the verbal exchange between speakers. & \textit{pointing (3)}, \textit{showing measurements (4)} \\ \addlinespace
Regulators & Regulators control the flow and pacing of conversation, helping to signal turn-taking or conversational shifts. For instance, ``nodding'' indicates agreement, while ``drawing circles in the air'' suggests speeding up, and ``holding palms out'' signals a desire to pause. & \textit{nodding (5)}, \textit{drawing circles in the air (6)}, \textit{holding palms out (7)} \\ \addlinespace
Adaptors & Adaptors are unconscious, self-directed movements that help individuals manage internal emotional states or satisfy personal needs. These gestures often arise in moments of stress or discomfort \cite{neff2011don}. & \textit{twirling or scratching hair (8)} \\ \addlinespace
Affect displays & Affect displays are gestures that reveal a person’s emotional state, often without accompanying speech. These include both spontaneous expressions and socially conditioned emotional cues. & \textit{laughing (9)}, \textit{arm crossing (10)}, \textit{hugging (11)} \\ \addlinespace
\bottomrule
\end{tabular}
    \begin{tablenotes}
      \footnotesize
      \item  \textit{Notes}: The numbers in parentheses represent each interaction's activity label.
    \end{tablenotes} \vspace{-0.3cm}
\end{threeparttable}
\end{table*}


\vspace{24pt}
\noindent \uppercase{\textbf{Kinesics Recognition Framework}} \vspace{12pt}

\begin{figure}[!t] 
  \centering{\includegraphics[width=\columnwidth]{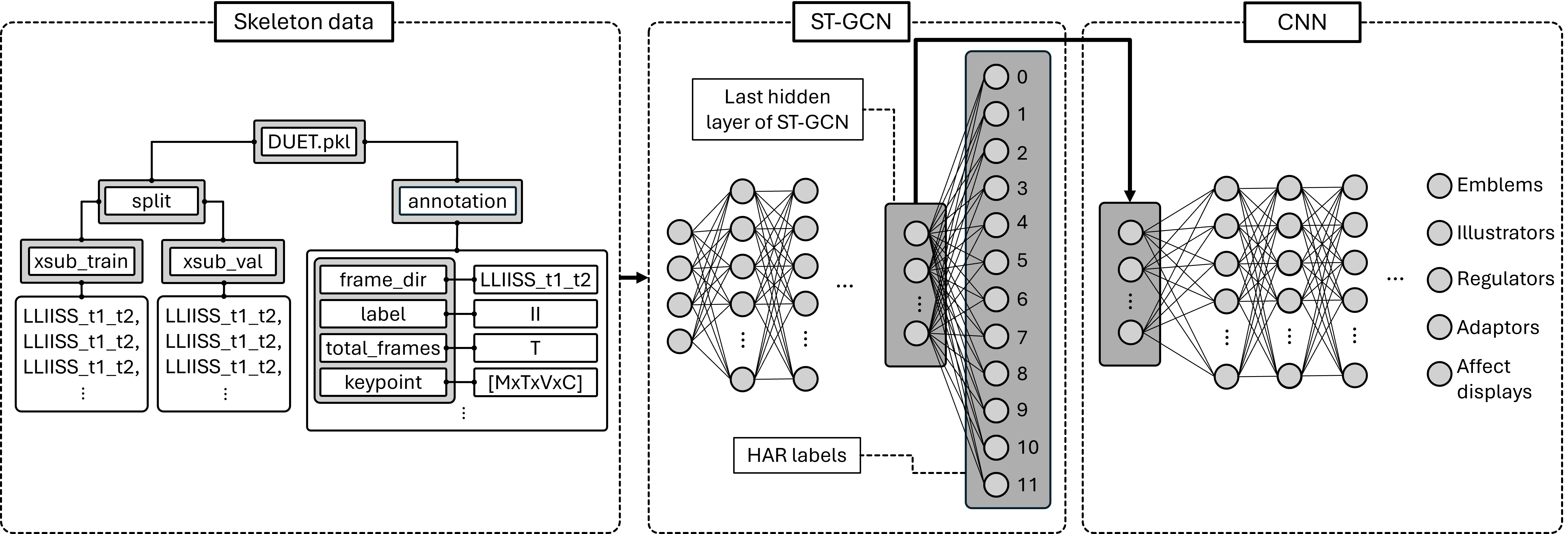}}
    \caption{\small The kinesics recognition framework comprises skeleton data preparation, ST-GCN, and CNN. (Note: In ``Skeleton Data,' double-bounded boxes represent dictionary keys, and each connecting line points to the corresponding value in the layer below.)
    }
\label{fig:flowchart}%
\end{figure}

Building on DUET's foundation for integrating HAR and psychological theory, we develop a framework for recognizing kinesic functions without relying on a predefined kinesic dictionary. As shown in Figure \ref{fig:flowchart}, the framework consists of three components: (1) skeleton joint data preparation, (2) ST-GCN, and (3) CNN for kinesics recognition. The full codebase and model architecture are publicly available in the DUET kinesics recognition repository \cite{duetgithub}. In this section, we describe each component of the framework in detail and present the results of its implementation. All notations introduced are used consistently throughout the section.

\vspace{12pt}
\noindent \textbf{Skeleton Data Preparation} \vspace{12pt}

The first step of the kinesics recognition framework is to process the data so that it is aligned with the ST-GCN data format. In DUET, one skeleton joint data sample is captured as a \texttt{csv} file, which stores an array of shape \( T\times (M\times V\times C)\). \(T\), \(M\), \(V\), and \(C \) stand for the number of frames, the number of subjects (i.e., 2), the number of keypoints (i.e., 32), and dimensions (i.e., $x$, $y$, and $z$ coordinates), respectively. This format is not compatible with that of ST-GCN. In fact, ST-GCN calls for an assembly of skeleton keypoints and their metadata to be condensed in a nested Python dictionary, which is serialized as a pickle (\texttt{pkl}) file as displayed in ``Skeleton data'' in Figure \ref{fig:flowchart}. The \texttt{DUET.pkl} file contains two nested dictionaries: \texttt{split} and \texttt{annotation}. The \texttt{split} dictionary specifies the training and validation partitions for ST-GCN and stores them in the lists \texttt{xsub\_train} and \texttt{xsub\_value}, respectively. Every element in the lists is the sample name in the form \texttt{LLIISS\_t1\_t2}. Here, \texttt{LL} stands for the location, which can be \texttt{CC} (a confined indoor space), \texttt{CM} (an open indoor space), or \texttt{CL} (an open outdoor space). \texttt{II} denotes numbers ranging from 0-11, which are the enumeration of interactions listed in Section 2. \texttt{SS} identifies the subject pairs ranging from 1-10. Last, \texttt{t1} and \texttt{t2} are the start and end timestamps of the interaction. The annotation dictionary stores a list of samples, each containing skeleton data along with associated metadata. Each sample includes the sample name (\texttt{frame\_dir}), activity class label (\texttt{label}), total number of frames (\texttt{total\_frames}), and skeleton keypoints (\texttt{keypoint}). A notable detail is that we extract only 25 of the 32 skeleton joints provided in DUET, as ST-GCN operates on a reduced set of keypoints. For our experiments, we perform cross-subject evaluation by designating participants \texttt{CCII01} and \texttt{CMII10} as the test set, with the remaining data used for training. This split is applied to both the ST-GCN and CNN components. Once the data are compiled in the required format, they are fed into the ST-GCN model to capture the structural patterns encoded in the skeleton keypoints.

\vspace{12pt}
\noindent \textbf{Kinesics Recognition} \vspace{12pt}

Human movements can appear unpredictable---let alone when represented as skeleton keypoint data---making it challenging to identify patterns that convey their underlying kinesic functions.  To recognize the kinesic functions of human activities, we construct a transfer learning model that extracts intrinsic patterns from skeleton data to infer the communicative purpose of each activity. As illustrated as the ``ST-GCN'' and ``CNN'' layers in Figure \ref{fig:flowchart}, the model consists of ST-GCN as fixed layers and CNN as modifiable layers. In this model, ST-GCN is not used in its conventional HAR role for classifying activity types. Instead, it serves to compress the high-dimensional skeleton data into a more compact representation that preserves essential activity features. We extract the output from ST-GCN's final hidden layer and use it as the input to the CNN. Leveraging the CNN's pattern recognition capabilities and the condensed keypoint representation, the model learns to distinguish the structural features that define each kinesic function.

\vspace{12pt}
\noindent \textbf{Experiment on DUET} \vspace{12pt}

To evaluate the performance of the kinesics recognition framework, we implemented the framework on five different subsets of DUET, as shown in Table \ref{tab:results}. This design choice was motivated by the initial underperformance of ST-GCN. When applied to the full set of 12 activities, ST-GCN struggled to accurately classify the activities, which in turn limited its ability to generate meaningful low-dimensional representations. These weak representations negatively impacted the downstream CNN, which failed to reliably recognize the underlying kinesic patterns in the skeleton data.

To address this issue, we constructed four additional subsets of DUET by selecting activities with more distinct movement patterns---thus maximizing the likelihood of improved ST-GCN accuracy. As shown in Table \ref{tab:results}, ST-GCN performance improves as the number of activity types decreases. With more accurate feature compression in ST-GCN, the final hidden layer can then be more effectively used as input to the CNN for kinesics classification. In general, the trends of ST-GCN and CNN performances align with each other---CNN improves when ST-GCN improves.

The results in Table \ref{tab:results} demonstrate a clear parallel between the performance of ST-GCN and CNN: the CNN is able to more accurately extract the kinesics of human activities when ST-GCN effectively encodes the distinguishing characteristics of each activity into a lower-dimensions representation. This finding has two key implications. First, improving the performance of ST-GCN alone can lead to better results for both HAR and kinesics classification. However, current limitations of ST-GCN hinder its ability to accurately classify all activities in DUET, particularly those involving subtle hand gestures. These fine-grained movements are difficult to capture with skeletal data, as each hand is represented by only three joints---fingertip, thumb, and wrist \cite{yan2018spatial}. If future developments enable ST-GCN to reliably classify these nuanced interactions, we anticipate that the corresponding kinesic functions will also become more accurately identifiable. Second, the observed alignment between ST-GCN and CNN performance suggests that there is an underlying structure within the skeleton data that governs the kinesic function of movements. To the best of our knowledge, this latent structure has not yet been systematically explored and presents a promising direction for future research.

\vspace{24pt}

\noindent \uppercase{\textbf{Conclusions and Future work}} \vspace{12pt} 

We introduce a kinesics recognition framework that uses transfer learning to infer the communicative functions of human activities based on a psychologically grounded taxonomy. By integrating ST-GCN with a CNN, the framework extracts latent structures from 3D skeleton joint data that reflect the kinesic categories---emblems, illustrators, regulators, adaptors, and affect displays. Traditionally, recognizing kinesic functions has relied on a one-to-one mapping between activities and categories, a method that lacks generalizability and incurs significant time and cost. Our approach eliminates the need for this manual mapping by uncovering intrinsic patterns within the data that govern kinesic expression. This not only improves accuracy and scalability, but also provides a richer, real-time input source for reinforcement learning models simulating human-environment interactions. Importantly, the use of anonymous 3D skeleton data preserves user privacy, aligning with ethical considerations central to human-centered research.

\begin{table}[!t]
    \centering
    \caption{\small Results of five subsets of DUET tested on the kinesics recognition framework.}
    \vspace{-6pt}
    \small
    \begin{tabular}{ c c c c }
    \toprule
    \makecell{Number of\\interactions} & Activity labels & \makecell{ST-GCN\\Accuracy (\%)} & \makecell{CNN\\Accuracy (\%)}\\ 
    \midrule
    4 & 2, 4, 8, 11 & 77 & 85 \\  
    6 & 2, 4, 5, 7, 8, 11 & 75 & 81 \\
    8 & 0, 2, 4, 6, 7, 8, 9, 11 & 70 & 70 \\  
    10 & 0, 1, 2, 3, 5, 6, 7, 8, 10, 11 & 67 & 58 \\
    12 & 0, 1, 2, 3, 4, 5, 6, 7, 8, 9, 10, 11 & 55 & 48 \\
    \bottomrule
    \end{tabular}
    \label{tab:results}
\end{table}

Future work will focus on establishing the functional relationship between the performances of ST-GCN and CNN. While our experimental results suggest a consistent trend between the two, a rigorous statistical analysis is needed to quantify this relationship---specifically, to assess its linearity and correlation. Establishing this link will enable the joint refinement of both components, advancing the framework's ability to translate human activities into deeper representations of psychological and cognitive states.

\vspace{24pt}

\noindent \uppercase{\textbf{Acknowledgements}} \vspace{12pt} 

This work is supported by the National Science Foundation under Grant \#2425121.

\vspace{24pt}

\small 

\bibliographystyle{iwshm}
\bibliography{IWSHM}


\end{document}